\documentclass[11pt]{article}
\usepackage{float}
\PassOptionsToPackage{table}{xcolor}

\usepackage[preprint]{acl}

\usepackage{times}
\usepackage{latexsym}
\usepackage[T1]{fontenc}
\usepackage[utf8]{inputenc}
\usepackage{microtype}
\usepackage{inconsolata}
\usepackage{graphicx}
\usepackage{capt-of}

\usepackage{amsmath}
\usepackage{amssymb}
\usepackage{url}

\usepackage{booktabs}
\usepackage{multirow}
\usepackage{array}
\usepackage{xspace}
\usepackage{graphicx}
\usepackage{subcaption}
\usepackage[table]{xcolor}
\usepackage{pifont}

\usepackage{hyperref}
\usepackage{fontawesome5}

\newcolumntype{M}{>{\raggedright\arraybackslash}p{3.4cm}}

\title{Hybrid Self-evolving Structured Memory for GUI Agents}

\author{
  Sibo Zhu\textsuperscript{1}\thanks{Equal contribution.} \quad
  Wenyi Wu\textsuperscript{1}\footnotemark[1] \quad
  Kun Zhou\textsuperscript{1}\thanks{Corresponding author.} \quad
  Stephen Wang\textsuperscript{2} \quad
  Biwei Huang\textsuperscript{1} \\
  \textsuperscript{1}University of California, San Diego \\
  \textsuperscript{2}Abel.ai \\
  \texttt{\{siz030, wew058\}@ucsd.edu} \quad
  \texttt{franciskunzhou@gmail.com}
}

\usepackage{enumitem}        
\usepackage{listings}        
\usepackage[most]{tcolorbox} 

\definecolor{DeepBlue}{HTML}{2B3298}
\definecolor{LightBlueBg}{HTML}{F3F6FF}
\definecolor{DeepGreen}{HTML}{2E7D32}
\definecolor{LightGreenBg}{HTML}{F1F8E9}
\definecolor{DeepOrange}{HTML}{EF6C00}
\definecolor{LightOrangeBg}{HTML}{FFF3E0}
\definecolor{DeepPurple}{HTML}{7B1FA2}
\definecolor{LightPurpleBg}{HTML}{F3E5F5}

\lstdefinestyle{promptstyle}{
  basicstyle=\ttfamily\footnotesize,
  columns=fullflexible,
  breaklines=true,
  keepspaces=true,
  showstringspaces=false,
  frame=none,
  backgroundcolor={}
}

\newtcolorbox{PromptBox}[2][]{%
  enhanced, breakable,
  colframe=DeepBlue,        
  colback=LightBlueBg,      
  coltitle=white,
  boxrule=0.6pt, arc=1.2mm,
  left=6pt,right=6pt,top=6pt,bottom=6pt,
  fonttitle=\bfseries,
  title={#2},
  #1                        
}

\newtcolorbox{PromptBoxFloat}[2][]{%
  enhanced,
  colframe=DeepBlue,        
  colback=LightBlueBg,      
  coltitle=white,
  boxrule=0.6pt, arc=1.2mm,
  left=6pt,right=6pt,top=4pt,bottom=4pt,
  fonttitle=\bfseries,
  title={#2},
  #1                        
}

\newcommand{\PromptSection}[2][DeepBlue]{%
  \par\medskip
  \begin{tcolorbox}[
    colback=#1!8!white,    
    colframe=#1!0,         
    left=4pt,right=4pt,top=2pt,bottom=2pt,
    arc=1mm, boxrule=0pt
  ]\centering\textbf{\textcolor{#1}{#2}}\end{tcolorbox}
  \vspace{-2pt}
}

\begin{document}
\maketitle

\begin{abstract}
The remarkable progress of vision–language models (VLMs) has enabled GUI agents to interact with computers in a human-like manner. Yet real-world computer-use tasks remain difficult due to long-horizon workflows, diverse interfaces, and frequent intermediate errors. Prior work equips agents with external memory built from large collections of trajectories, but relies on flat retrieval over discrete summaries or continuous embeddings, falling short of the structured organization and self-evolving characteristics of human memory. Inspired by the brain, we propose \textbf{Hybrid Self-evolving Structured Memory (\textsc{HyMEM})}, a graph-based memory that couples discrete high-level symbolic nodes with continuous trajectory embeddings. \textsc{HyMEM} maintains a graph structure to support multi-hop retrieval, self-evolution via node update operations, and on-the-fly working-memory refreshing during inference. Extensive experiments show that \textsc{HyMEM} consistently improves open-source GUI agents, enabling 7B/8B backbones to match or surpass strong closed-source models; notably, it boosts Qwen2.5-VL-7B by +22.5\% and outperforms Gemini2.5-Pro-Vision and GPT-4o.

\vspace{0.8em}
\noindent\hfil
\href{https://github.com/NickSiboZhu/HyMEM-GUI-Agent}{\faGithub~\textbf{Code}} \hspace{2em}
\href{https://huggingface.co/spaces/Nick0907/hymem-project-page}{\faGlobe~\textbf{Website}}
\hfil

\end{abstract}

\section{Introduction}
Benefiting from the strong perception and planning capabilities of large vision–language models (VLMs), VLM-based graphical user interface (GUI) agents have emerged as a promising direction for automating interactions on desktop and mobile platforms. To follow user instructions, these agents should interpret visual and linguistic context, reason and plan over dynamic layouts, predict and execute precise actions (\emph{e.g.,} clicking and typing)~\citep{qin2025uitarspioneeringautomatedgui, hong2023cogagent}. However, the long-horizon and highly diverse nature of real-world computer-use tasks still makes human-level performance challenging: current GUI agents often fail due to non-trivial intermediate errors or overlooking critical conditions~\cite{yang2025probenchbenchmarkingguiagents}.

To mitigate these limitations, recent work draws inspiration from the human brain’s memory mechanisms and equips GUI agents with external memory modules~\citep{wu2025autoscale_gui_memory, zhang2023appagentmultimodalagentssmartphone}. Humans can automatically store, organize, and update experiences—whether observed from others or acquired firsthand—and retrieve relevant memories as references when needed~\cite{williams2022power}. Similarly, existing approaches collect an agent’s past trajectories into a database (\emph{i.e.,} an experience memory), where each item’s multimodal content is either distilled into discrete tokens (\emph{e.g.,} sentences or key phrases)~\cite{ouyang2025reasoningbank} or compressed into continuous embeddings~\cite{wu2025comem}. At inference time, the agent retrieves relevant items via similarity matching and uses them to construct complementary context as a working memory for the current instruction.

Despite these advances, current agent memory designs remain far weaker than human memory, particularly in how they organize knowledge and evolve it over time. The brain can preserve fine-grained evidence for reconstructing multimodal episodes, associate them with higher-level concepts or strategies for efficient search and reasoning, and continuously update as new experiences arrive. Neurobiological evidence suggests that human memory operates as a hybrid system: the hippocampus encodes rich multimodal experiences into continuous latent representations~\citep{spens2024consolidation}, while the neocortex extracts statistical regularities and forms higher-level discrete symbols~\citep{mcclelland1995why}. New experiences are then linked to existing knowledge and only the novel information needs to be stored~\cite{Spens2024}. This structured organization supports long-term retention of massive multimodal details while enabling efficient retrieval, reasoning, and continual evolution.

\newcommand{\cmark}{\ding{51}}  
\newcommand{\xmark}{\ding{55}}  

\begin{table}[t]
\centering
\renewcommand{\arraystretch}{1.15}
\setlength{\tabcolsep}{3pt} 
\resizebox{\columnwidth}{!}{%
\begin{tabular}{l c c c c c}
\toprule

Method & Data Form & MM & Struct. & Global Up. & Local Up. \\
\midrule
MemGPT & Discrete & \xmark & \xmark & \xmark & \cmark \\

ExpeL & Discrete & \xmark & \xmark & \cmark & \xmark \\

ReasoningBank & Discrete & \xmark & \xmark & \cmark & \xmark \\

AWM & Discrete & \xmark & \xmark & \cmark & \xmark \\

A-MEM & Discrete & \xmark & \cmark & \cmark & \xmark \\

G-Memory & Discrete & \xmark & \cmark & \cmark & \xmark \\

CoMEM & Continuous & \cmark & \xmark & \xmark & \xmark \\

\midrule

\textbf{\textsc{HyMEM}} & \textbf{Hybrid} & \textbf{\cmark} & \textbf{\cmark} & \textbf{\cmark} & \textbf{\cmark} \\
\bottomrule
\end{tabular}%
}

\caption{Comparison of Memory Paradigms. Abbreviations: \textbf{MM}: Multimodal support; \textbf{Struct.}: Structured organization; \textbf{Up.}: Update. Methods are compared by data form, MM, and structure, as well as Global (self-evolution) and Local (working memory refresh) updates.}

\label{tab:memory_comparison}
\end{table}

Inspired by the human brain, we aim to build a hybrid structured external memory that organizes knowledge more effectively and supports flexible updates and self-evolution over time. Concretely, we first construct a hybrid graph-based memory from agent trajectories by merging trajectories into higher-level nodes with discrete symbolic semantics, while continuous embeddings of the trajectory are used to preserve fine-grained multimodal evidence. Building on this structure, we design a graph evolution and update mechanism that can add, update, and replace nodes as new trajectories arrive, enabling the memory to accumulate knowledge without uncontrolled growth. During inference, we further introduce an on-the-fly working-memory update that detects execution phase shifts and triggers re-retrieval and refreshing, allowing multi-turn interaction between the agent and the memory database throughout long-horizon tasks.

Based on these designs, we propose \textbf{Hybrid Self-evolving Structured Memory (\textsc{HyMEM})}, a graph-based external memory that tightly couples symbolic guidance with multimodal experience to improve planning and grounding for GUI agents. Empirically, \textsc{HyMEM} consistently boosts open-source GUI agents across multiple benchmarks, and can enable lightweight 7B/8B-scale backbones to match or surpass strong closed-source models. For example, it improves Qwen2.5-VL 7B~\cite{bai2025qwen25vltechnicalreport} by +22.5\% (12.5\% to 35.0\%), exceeding Gemini2.5-Pro-Vision~\cite{team2023gemini} by 5.4\%, and GPT-4o~\cite{openai2024gpt4o} by 15.3\%. Similar gains are observed on UI-TARS-1.5-7B~\cite{qin2025uitarspioneeringautomatedgui} and Qwen3-VL-8B~\cite{bai2025qwen3vltechnicalreport}, demonstrating \textsc{HyMEM} as a cost-effective path to state-of-the-art GUI agents.

\section{Related Work}
\label{sec:related_work}

\paragraph{GUI Agents.}
Early GUI agents primarily relied on HTML or DOM parsing to ground actions~\citep{mind2web, zheng2024seeact_multimodal_mind2web}, but the field has rapidly shifted toward Vision-Language Models (VLMs) capable of perceiving screens visually~\citep{hong2023cogagent, qin2025uitarspioneeringautomatedgui}. While autonomous frameworks like AppAgent~\citep{zhang2023appagentmultimodalagentssmartphone} integrate these vision experts into closed-loop systems, maintaining consistency across long-horizon, open-ended tasks remains a significant challenge. This bottleneck has driven recent research toward equipping agents with external memory systems~\cite{hu2025memoryageaiagents}, allowing them to retain long-term context and generalize from historical interactions rather than relying solely on immediate perception~\cite{hu-etal-2025-hiagent}.

\paragraph{Language Model Memory.}
Current memory mechanisms generally fall into discrete, continuous, or graph-based paradigms. Discrete approaches log experiences as textual summaries~\citep{packer2024memgptllmsoperatingsystems} or reasoning traces~\citep{ouyang2025reasoningbank}; while interpretable, they often fail to capture the fine-grained visual nuances of GUI elements~\cite{luo2025visualtesttimescalinggui}. Conversely, continuous memories compress multimodal inputs into dense embeddings~\citep{wu2025comem, wu2025autoscale_gui_memory} or latent tokens~\citep{zhang2025memgenweavinggenerativelatent}, preserving sensory details but creating an information bottleneck that hinders explicit reasoning. To improve organization, graph-based systems~\citep{xu2025amem, zhang2025gmemorytracinghierarchicalmemory} structure knowledge as connected nodes, enabling agents to perform multi-hop retrieval and reasoning beyond similarity matching~\citep{shen-etal-2025-gear}.

Hybrid architectures attempt to merge these formats to balance abstraction with detail. However, existing hybrid methods like ExpeL~\citep{zhao2024expelllmagentsexperiential} often remain purely token-based, pairing trajectories with textual insights but potentially missing visual fidelity. In contrast, our \textbf{\textsc{HyMEM}} organizes discrete textual insights (capturing high-level strategies) and continuous embeddings (preserving fine-grained multimodal details) within a graph structure, ensuring the agent possesses both strategic abstraction and precise perceptual grounding.

\section{Methodology}
\label{sec:methodology}

\begin{figure*}[t]
  \centering
  \includegraphics[width=\textwidth]{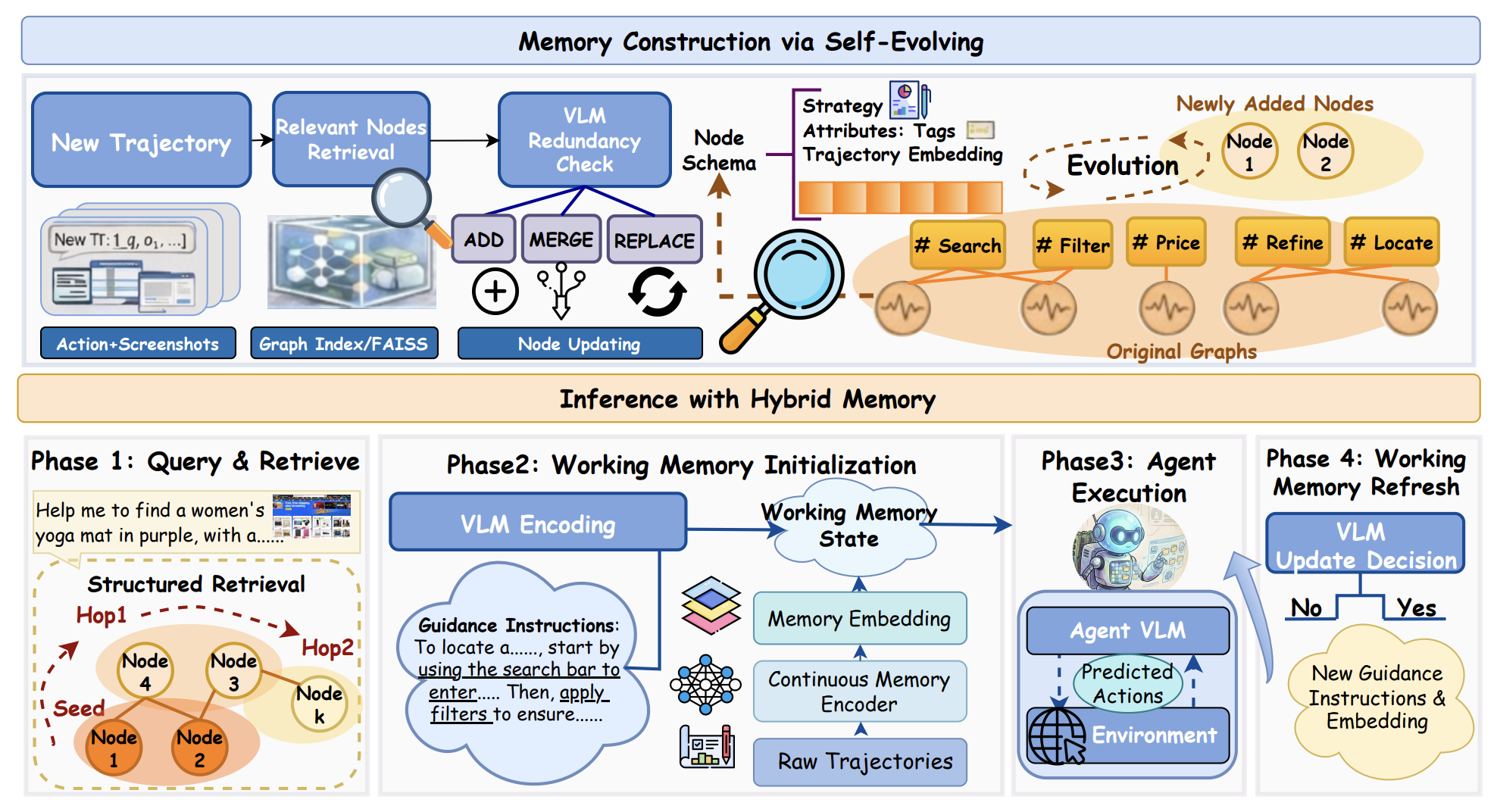}
  \caption{Overview of the \textbf{Hybrid Self-Evolving Memory (\textsc{HyMEM})} system. Top: memory construction via graph evolution, where new trajectories are integrated using retrieval and VLM-based redundancy checks. Bottom: four inference phases—(1) \textit{Query \& Retrieve} via structured graph expansion, (2) \textit{Working Memory Initialization} using hybrid encoding, (3) \textit{Agent Execution}, and (4) \textit{On-the-fly Working Memory Refresh} based on new observations.}
  \label{fig:overview}
\end{figure*}

We propose \textbf{Hybrid Self-evolving Structured Memory (\textsc{HyMEM})}, a graph-based external memory for guiding GUI agents. It is a brain-inspired hybrid memory system with a hippocampal-like Continuous Pathway that encodes trajectories as latent embeddings and a neocortical-like Discrete Pathway that extracts symbolic schemas for high-level concepts. 
Based on that, \textsc{HyMEM} organizes the memory into an evolving graph, enabling structured search to gather diverse related experiences, and continuously self-evolution for global updates and on-the-fly local refresh during execution.

\subsection{Hybrid Structured Memory}
\label{sec:memory_construction}

\paragraph{Graph Schema.}
We maintain the memory as an evolving graph \(G=(V,E)\), where $V$ and $E$ denote the set of nodes and edges. 

The node set \(V\) consists of Trajectory Nodes, where each node \(v_i \in V\) represents a specific, successful interaction sequence. To support both high-level planning and low-level execution, we define each node as a tuple \(v_i = (c_i, A_i, m_i)\):

$\bullet$ High-level Strategy (\(c_i\)): A heuristic summarizing the core strategy (\emph{e.g.,} price filter low-to-high).

$\bullet$ Middle-level Attributes (\(A_i\)): A semantic tag (\emph{e.g.,} \texttt{\#search}, \texttt{\#filter}, \texttt{\$price}) providing cues about actions, UI elements, and domain concepts.

$\bullet$ Low-level Trajectory Embeddings (\(m_i\)): The continuous representations of the trajectory, to preserve fine-grained multimodal details.

We establish undirected edges between trajectory nodes that share identical Middle-level Attributes. This design yields an associative topology that supports multi-hop structured search.

\paragraph{Node Encoding.}
To guarantee the efficiency and low cost, we leverage lightweight VLMs (\emph{i.e.,} Qwen2.5-VL-7B), for encoding the three types of nodes.
For trajectory embedding nodes, we use CoMEM~\cite{wu2025comem,wu2025comemagent} to condense the whole trajectory into 8 embeddings that can be directly concatenated into GUI agent embedding layers for reading.
For strategy nodes, we prompt a VLM to summarize a concise heuristic describing the high-level strategy behind the trajectory, based on its task description and action sequence.
For attribute nodes, we then extract word- and phrase-level semantic tags from the generated strategy using the same VLM.
In this way, the strategy and attribute node representations are discrete word tokens, and the trajectory nodes are continuous embeddings.

\subsection{Memory Construction via Self-Evolving}
\label{sec:self_evolving_construction}

We construct the memory in a self-evolving manner: as new trajectories arrive, the memory graph is incrementally refined to add new nodes and update existing structures. To control memory quality, we adopt a three-stage pipeline for adding each new trajectory: retrieving relevant nodes, checking redundancy via information gain, and applying structured update.

\paragraph{Relevant Nodes Retrieval.}
Given a new trajectory $\tau=[q,o_1,a_1,\cdots]$, we retrieve a small set of related nodes to serve as the comparison context for subsequent updates. Concretely, we construct the semantic vector for each trajectory by concatenating the normalized CLIP representations of the text query $q$ and the first observation $o_1$:
\begin{equation}
    \mathbf{v} = [\text{CLIP}_{txt}(q); \text{CLIP}_{img}(o_1)] \label{eq-match}
\end{equation}
where $\text{CLIP}_{txt}$ and $\text{CLIP}_{img}$ denote the text and image encoders of CLIP~\cite{radford2021learningtransferablevisualmodels}. Then, we use FAISS~\cite{johnson2017billionscalesimilaritysearchgpus} to perform nearest-neighbor search based on cosine similarity to obtain the top-$K$ most similar trajectory nodes (and their linked attribute/strategy nodes). 

\paragraph{Redundancy Checking.}
Similarity cannot distinguish whether a new trajectory provides sufficient information gain in current memory. We therefore introduce a redundancy check using a VLM judge. 

Conditioned on the new trajectory and the retrieved neighbors (along with their first screenshots for visual context), the judge evaluates: (i) whether the new trajectory represents a genuinely different strategy not covered by existing nodes (\textsc{Add}), (ii) whether it describes the same strategy as an existing node but can improve the existing takeaway by incorporating complementary insights (\textsc{Merge}), or (iii) whether it covers the same strategy but is strictly superior in specificity and actionability, warranting replacement of the existing node (\textsc{Replace}). This decision explicitly measures the \emph{marginal utility} of storing the trajectory and drives the subsequent structured update operation.

\paragraph{Memory Update.}
Based on the judge results, we update the graph by following operations:

$\bullet$ ADD: If the trajectory introduces new attributes or a previously unseen strategy, we create the corresponding strategy and/or attribute nodes (if missing), insert the new trajectory embedding node, and connect it to its attributes and strategy.

$\bullet$ MERGE: If the trajectory matches an existing strategy but provides complementary evidence (\emph{e.g.,} a new UI variant or a different action path), we attach the new trajectory node to the matched attribute nodes and update the associated strategy node by consolidating newly observed attributes.

$\bullet$ REPLACE: If the trajectory is a strictly better instance of an existing strategy (\emph{e.g.,} fewer steps, higher success, or fewer risky actions under the same conditions), we replace the outdated trajectory evidence and refresh the attribute connections to reflect the improved execution.

After node updates, we evolve the graph connectivity by instantiating or strengthening edges between strategy and attribute nodes according to newly observed co-occurrences. This continual refinement encourages stable regularities to emerge over time, making the memory increasingly coherent, less redundant, and easier to retrieve via associative multi-hop search.

\subsection{Memory Utilization}
\label{sec:memory_utilization}
Based on the constructed hybrid structured memory, we maintain an agent \emph{working memory} as dynamic inference-time context that guides execution and can be refreshed as the GUI state evolves. Concretely, we first retrieve relevant experiences from the global memory to initialize the working memory, and then update it on-the-fly whenever the task transitions into a new interaction phase.

\paragraph{Structured Retrieval.}
\label{method:retrieval}
We start by identifying a set of high-relevance \emph{seed} nodes via semantic matching (Eq.~\ref{eq-match}). Specifically, we perform $k$-nearest neighbor search with the multimodal query embedding $\mathbf{v}$ against the stored node embeddings to obtain an initial candidate set $\tilde{N}$. To improve coverage beyond surface-level similarity, we expand retrieval through the graph: for each seed node $n \in \tilde{N}$, we collect its 1-hop neighbors, re-rank them by cosine similarity to $\mathbf{v}$, and add the top-$k$ neighbors into $\tilde{N}$. Since these neighbors often capture conceptually necessary knowledge that may not be visually similar to the current screen, we repeat this expand-and-rerank procedure for $t$ iterations, resulting in a diverse yet relevant set of nodes for constructing the working memory.

\paragraph{Working Memory Initialization.}
Given the retrieved node set $\tilde{N}$, we form the working memory in two complementary views. First, we digest the associated strategy and attribute nodes into concise \emph{Guidance Instructions} using a VLM, which filters irrelevant details and synthesizes actionable, strategy-level advice; these instructions are injected into the system prompt to steer high-level planning. Second, following CoMEM~\cite{wu2025comemagent,wu2025comem}, we concatenate the corresponding continuous trajectory embeddings with the VLM input to provide implicit access to fine-grained visual/action evidence. Together, the discrete Guidance Instructions act as a semantic anchor that primes attention, while the continuous embeddings preserve details that are difficult to express textually, enabling more grounded and reliable action prediction.

\paragraph{On-the-fly Working Memory Refresh.}
Because GUI tasks can change substantially after certain actions, a fixed retrieved context can quickly become stale. We therefore enable on-the-fly local evolution during agent inference. After each action, a VLM compares the transition $(o_t \rightarrow o_{t+1})$ with the current Guidance Instructions to detect phase shifts (\emph{e.g.,} from ``search'' to ``checkout''). When a shift is detected, the VLM further determines which parts of the current working memory should be preserved (\emph{e.g.,} long-term goals and constraints) and which should be discarded. We then rerun retrieval conditioned on the updated state, merge newly retrieved guidance with the preserved takeaways, and refresh both the Guidance Instructions and the continuous embeddings. This cyclic update keeps the context compact and relevant, ensuring that the agent remains synchronized with the evolving GUI environment throughout long-horizon execution.

\section{Experiments}
\label{sec:experiments}

\subsection{Experiment Settings}
This section presents our experimental setup, including the agent's action interface, memory construction and retrieval settings, evaluation benchmarks, and baseline comparisons.

\label{sec:env setup}
\paragraph{Action Interface.}
Our agent follows a ReAct-style paradigm, combining structured reasoning with tool-based execution. The model accepts screenshot multimodal inputs and interacts with a predefined toolset comprising GUI tools (\emph{e.g.,} \texttt{CLICK}, \texttt{TYPE}, \texttt{SCROLL PAGE}, \texttt{WAIT}, \texttt{GO BACK}, \texttt{STOP}) and semantic tools (\emph{e.g.,} \texttt{Page Content Analyzer}, \texttt{Change Page}).
To support accurate grounding (\emph{e.g.,} clicking or typing on specific UI elements), we adopt a two-stage approach:
\begin{itemize}
    \item \textit{Stage 1:} Screenshots are augmented using a SOM-based method, which assigns unique labels to UI elements. The agent generates references based on these labels.
    \item \textit{Stage 2:} If the agent fails to produce a valid label, we fall back to UI-INS-7B~\cite{chen2025uiinsenhancingguigrounding}, a robust vision-language grounding model, to resolve the location.
\end{itemize}
This hybrid mechanism ensures robust and reliable UI interaction, especially in visually complex or dynamic web environments.

\paragraph{Memory Settings.}
\label{sec:appendix_memory_setting}
We collect a total of 2,883 successful trajectories from the following sources: GUIAct~\cite{guiact}: A curated set of successful agent behaviors; Mind2Web Training Set~\cite{mind2web, zheng2024seeact_multimodal_mind2web}: Human-annotated demonstrations; Agent Rollouts: High-confidence trajectories generated by our own agent during training. In the impact of memory size ablation study \ref{sec:memory_size}, we also include trajectories from the automated data flywheel\cite{wu2025comemagent} to expand the memory graph.
To support structured and scalable retrieval, we organize the trajectories into a graph: Each node represents one or more similar trajectories, annotated with metadata such as \textit{task descriptions}, \textit{strategy}, and \textit{attributes}. Edges are created between nodes that share overlapping tags. The final graph we used in the main table \ref{tab:main_results} contains 1,858 nodes and over 1 million edges. Some trajectories are merged into the same node for redundancy reduction.
At inference time, the agent retrieves 10 relevant trajectories with the structured retrieval method introduced in \ref{method:retrieval}: Top-5 trajectories are selected as seed nodes based on multimodal (visual + text) similarity to the current input. The remaining 5 are selected via graph expansion by taking 1-hop neighbors of the seed nodes.




\paragraph{Training Recipes.}
we perform parameter-efficient fine-tuning on the Q-Former and LoRA layers within the VLM memory encoder, and only 1.2\% of the total model parameters are updated during training. Empirically, we observe that training converges rapidly, and a single epoch is sufficient to achieve strong performance. More details about training settings can be found in \ref{appendix:training_recipe}.

\paragraph{Evaluation Benchmarks.}
We evaluate our approach on three challenging multimodal web-agent benchmarks: WebVoyager~\citep{he2024webvoyager}, Multimodal-Mind2Web~\citep{mind2web, zheng2024seeact_multimodal_mind2web}, and MMInA~\citep{tian2025mminabenchmarkingmultihopmultimodal}.

For evaluation on WebVoyager and Multimodal-Mind2Web, we adopt the LLM-as-Judge protocol~\citep{he2024webvoyager}, where we feed the task description and screenshot trajectory to a VLM to determine if the task was successfully completed. For MMInA, we compare the model's final answer with the ground truth using a language model. We report Task Accuracy as the primary metric.

\medskip
\noindent To assess the effectiveness of our proposed memory-augmented framework, we compare the performance against the following baselines:

\textbf{Closed-Source Base Models:} GPT-4o~\citep{openai2024gpt4o}, Gemini2.5-Pro-Vision~\citep{team2023gemini}, and Claude-4~\citep{anthropic2024claude4}, to provide a comparison with strong proprietary base model capabilities.

\textbf{Open-Source Base Models:} We select a diverse set of open-source models to evaluate performance across different scales and adaptations. This includes large-scale models like Qwen2.5-VL-32B~\citep{bai2025qwen25vltechnicalreport}, and task-specific models such as CogAgent~\citep{hong2023cogagent} and WebSight~\citep{bhathal2025websight}. Additionally, we evaluate on smaller LLMs including Qwen2.5-VL-7B~\citep{bai2025qwen25vltechnicalreport}, Qwen3-VL-8B~\citep{bai2025qwen3vltechnicalreport}, and UI-TARS-1.5-7B~\citep{qin2025uitarspioneeringautomatedgui}, which serve as the foundation for our experiments.

\textbf{Open-Source + Memory-Augmented Models:}
\begin{itemize}
    \item \textit{Textual Memory:} ReasoningBank~\citep{ouyang2025reasoningbank} and Agent Workflow Memory (AWM)~\citep{wang2024awm} are advanced text-based memory methods. They distill experience from past trajectories in the form of tokenized textual prompts, serving as a baseline to assess the impact of text-based memory.
    \item \textit{Discrete Memory:} Our graph-based retrieval system retrieves relevant trajectories and digests these experiences using a VLM to provide contextual guidance for the current state.
    \item \textit{Continuous Memory:} Following~\citep{wu2025comemagent}, we encode multimodal trajectory data into continuous embeddings, evaluating the benefits of rich, multimodal memory representations for complex agent tasks.
    \item \textit{Hybrid Memory:} A fusion of discrete and continuous memory that allows for dynamic memory updating during task execution.
\end{itemize}

All results in Table~\ref{tab:main_results} are based on our own implementation. We maintain consistent environment setup, task sampling, and evaluation criteria across all models to ensure fair comparisons.

\subsection{Result Analysis}
\label{sec:results}

\begin{table*}[h]
\centering
\renewcommand{\arraystretch}{1}
\setlength{\tabcolsep}{3pt}
{\small
\resizebox{\textwidth}{!}{%
\begin{tabular*}{\linewidth}{@{\extracolsep{\fill}} l l 
c c c c    
c c c c    
c          
c}         
\toprule

\multicolumn{2}{c}{} 
& \multicolumn{4}{c}{\textbf{WebVoyager}} 
& \multicolumn{4}{c}{\textbf{Mind2Web}} 
& \multicolumn{1}{c}{\textbf{MMInA}} 
& \multirow{2}{*}{\textbf{Overall}}
\\
\cmidrule(lr){3-6} \cmidrule(lr){7-10} \cmidrule(lr){11-11} 
\textbf{Backbone} & \textbf{Model/Method} 
& \textbf{Amz} & \textbf{Cour} & \textbf{Recp} & \textbf{Map} 
& \textbf{Info} & \textbf{Svc} & \textbf{Ent} & \textbf{Trav}
& \textbf{Wiki} &  \\
\midrule

\multirow{3}{*}{Closed-Source}
& GPT-4o              & 24.4\% & 7.1\%  & 13.3\% & 36.6\% & 7.8\%  & 14.1\% & 3.4\%  & 19.4\% & 51.0\% & 19.7\% \\
& Gemini-Pro-Vision   & 41.7\% & 42.9\% & 20.0\% & 53.7\% & 16.7\% & 22.4\% & 0.0\%  & 19.4\% & 50.0\% & 29.6\% \\
& Claude-4            & 63.4\% & 28.6\% & 33.3\% & 70.0\% & 25.6\% & 40.0\% & 6.9\%  & 9.7\%  & 52.0\% & 36.6\% \\
\midrule

\multirow{4}{*}{Open-Source}
& Qwen2.5-VL-32B      & 46.3\% & 26.2\% & 6.7\%  & 29.3\% & 14.1\% & 20.0\% & 6.9\%  & 9.7\%  & 43.0\% & 22.5\% \\  
& CogAgent            & 12.2\% & 9.5\%  & 26.7\% & 9.8\%  & 24.4\% & 8.2\%  & 13.8\% & 16.1\% & 21.0\% & 15.7\% \\
& Websight            & 24.4\% & 4.8\%  & 13.3\% & 29.3\% & 10.3\% & 3.5\%  & 3.4\%  & 0.0\%  & 12.0\% & 11.2\% \\
\midrule

\multirow{6}{*}{UI-TARS-1.5-7B}
& Baseline               & 31.7\% & 16.7\% & 20.0\% & 31.7\% & 6.4\%  & 4.7\%  & 6.9\%  & 0.0\%  & 36.0\% & 17.1\% \\
& + Reasoning Bank         & 0.0\% & 11.9\% & 6.7\% & 12.2\% & 7.7\% & 8.2\%  & 3.7\% & \textbf{6.7\%} & 44.0\% & 11.2\%\\
& + AWM                     & 4.9\% & 7.1\% & 4.4\% & 14.6\% & 10.3\% & 3.5\% & 3.7\% & 3.3\% & 26.0\% & 8.6\% \\
& + Discrete               & 19.5\% & 16.7\% & 8.9\% & 17.1\% & 11.5\% & 8.2\% & 7.4\% & 3.4\% & 46.0\% & 15.4\% \\
& + Continuous             & 43.9\% & \textbf{33.3\%} & 24.4\% & 43.9\% & 16.7\% & \textbf{28.2\%} & 10.3\% & 3.2\% & 54.0\% & 28.7\% \\
\rowcolor{cyan!8}
& + \textbf{HyMEM} & \textbf{58.5\%} & 28.6\% & \textbf{26.7\%} & \textbf{53.7\%} & \textbf{16.7\%} & 25.9\% & \textbf{10.3\%} & 6.5\% & \textbf{54.0\%} & \textbf{31.2\%} \\
\midrule

\multirow{6}{*}{Qwen2.5-VL-7B}
& Baseline               & 14.6\% & 2.4\%  & 15.9\% & 16.7\% & 9.0\%  & 11.8\% & 0.0\%  & 4.4\%  & 38.0\% & 12.5\% \\
& + Reasoning Bank         & 29.3\% & 9.5\% & 6.7\%  & 29.3\% & 9.0\% & 20.0\% & 3.4\% & 6.5\% & 44.0\% & 17.5\% \\
& + AWM                    & 17.1\% & 4.8\% & 11.1\% & 29.3\% & 7.7\% & 10.6\% & 0.0\% & 6.5\% & 31.0\% & 13.1\% \\
& + Discrete               & 22.0\% & 21.4\% & 8.9\%  & 31.7\% & 9.0\% & 12.9\% & 10.3\% & 3.2\%  & 34.0\% & 17.0\% \\
& + Continuous             & 24.4\% & 17.1\% & 8.9\%  & 34.1\% & 16.7\% & 23.5\% & \textbf{10.3\%} & 12.9\% & 47.0\% & 21.7\% \\
\rowcolor{cyan!8}
& + \textbf{HyMEM} & \textbf{63.4\%} & \textbf{54.8\%} & \textbf{20.0\%} & \textbf{53.7\%} & \textbf{17.9\%} & \textbf{23.5\%} & 3.4\% & \textbf{22.6\%} & \textbf{56.0\%} & \textbf{35.0\%} \\
\midrule

\multirow{6}{*}{Qwen3-VL-8B}
& Baseline               & 36.6\% & 16.7\% & 13.3\% & \textbf{43.9\%} & 9.0\%  & 20.0\% & 10.3\% & 9.7\%  & 38.0\% & 21.9\% \\
& + Reasoning Bank         & 36.6\% & 11.9\% & 17.8\% & 31.7\% & 10.3\% & \textbf{24.7\%} & \textbf{17.2\%} & 6.5\% & \textbf{48.0\%} & 22.7\% \\
& + AWM                    & 39.0\% & 19.0\% & 8.9\%  & 31.7\% & 12.5\% & 16.5\% & 6.9\%  & 16.1\% & 48.0\% & 22.1\% \\
& + Discrete               & 31.7\% & 16.7\% & 15.6\% & 31.7\% & 11.5\% & 16.5\% & 13.8\% & 16.1\% & 38.0\% & 21.3\% \\
& + Continuous             & 43.9\% & 19.0\% & 17.8\% & 34.1\% & 12.5\% & 17.9\% & 3.4\%  & 19.4\% & 42.0\% & 23.3\% \\
\rowcolor{cyan!8}
& + \textbf{HyMEM} & \textbf{46.3\%} & \textbf{19.0\%}    & \textbf{26.7\%}    & 39.0\% & \textbf{17.9\%}    & 20.0\%    & 6.9\%    & \textbf{25.8\%}    & 47.0\% & \textbf{27.6\%} \\

\bottomrule
\end{tabular*}
}
}
\caption{Task success rates (\%) across WebVoyager, Mind2Web, and MMInA. Bolded entries indicate best-performing open-source models in each domain. Overall is averaged across available domains.}
\label{tab:main_results}
\end{table*}

Table \ref{tab:main_results} compares task success rates across three benchmarks. 
The results  highlight the impact of different memory strategies across various base models. Notably, HyMEM consistently improves performance and even outperforms strong closed-source models like GPT-4o and Gemini2.5-Pro-Vision.
Without memory, open-source baselines such as Qwen2.5-VL-7B and Qwen3-VL-8B perform not well, achieving 12.5\% and 21.9\% respectively. Larger or specialized models like Qwen2.5-VL-32B, CogAgent, and WebSight also struggle, indicating that model scale or task-specific finetuning alone is insufficient for long-horizon web tasks.
Adding advanced textual memory summaries (\emph{e.g.,} Reasoning Bank, AWM) yields moderate gains (\emph{e.g.,} +5.0\%, +0.6\% on Qwen2.5-VL-7B), but these methods are limited by their unimodal nature and lack of memory management mechanisms like updating or pruning. Consequently, they often fail to capture the rich, evolving context required for robust reasoning in complex environments.
In contrast, both Discrete and Continuous memory individually offer consistent gains. For instance, Qwen2.5-VL-7B improves from 12.5\% to 17.0\% (Discrete) and 21.7\% (Continuous); Qwen3-VL-8B shows similar trends. However, neither method alone is universally effective. For example, UI-TARS-1.5-7B performs better with continuous memory than with discrete memory, highlighting the complementary strengths of them. 

The HyMEM approach, which combines discrete high level concepts with continuous multimodal representations, achieves the best results across all open-source models. Qwen2.5-VL-7B + Hybrid reaches 35.0\% (+22.5\%), outperforming Claude-4 in domains like Travel and Wikipedia, and surpasses  GPT-4o and Gemini2.5-Pro-Vision in average. Qwen3-VL-8B and UI-TARS-1.5-7B also achieve their highest scores with HyMEM. These findings demonstrate that Hybrid Memory enables agents to reason wisely in the detailed multimodal context, which is a key capability for solving complex and multi-step web tasks.

\subsection{Further Analysis}

\subsubsection{Memory Self-evolving Analysis}
\label{sec:self_evolving_memory_ablation}

\begin{figure}[h]
    \centering
    \includegraphics[width=\linewidth]{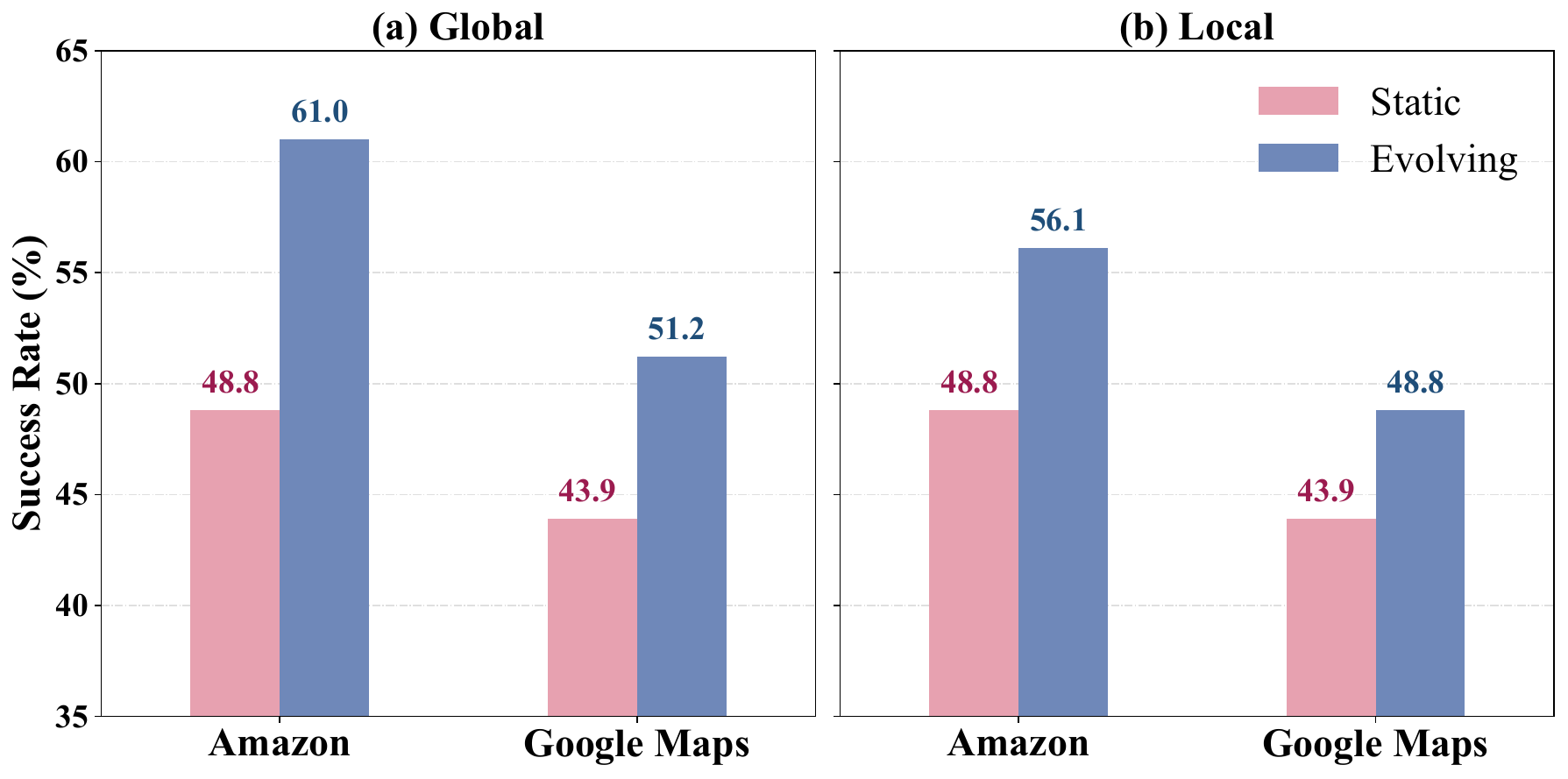}
    \caption{Success rate comparison between static memory and self-evolving \textsc{HyMEM}. (a): Global evolution. (b): Local evolution via working memory refresh.}
    \label{fig:self_evolve}
\end{figure}

We evaluate the self-evolving mechanisms of \textsc{HyMEM} by comparing a static-memory baseline with the self-evolving variant on Amazon and Google Maps domains. As shown in Figure~\ref{fig:self_evolve}, we isolate the contributions of \textit{global evolution}, the long-term consolidation, and \textit{local evolution}, the working memory adaptation.


\paragraph{Global evolution} drives continuous learning by integrating successful experiences into reusable strategies. As shown in Figure~\ref{fig:self_evolve} (a), this yields substantial relative gains: $\sim$25\% on Amazon (48.8\% $\to$ 61.0\%) and $\sim$16.6\% on Google Maps (43.9\% $\to$ 51.2\%). Our VLM-based deduplication balances growth with compression, where half of successful trajectories (\emph{e.g.,} 52\% on Google Maps) trigger UPDATE operations to merge existing strategies rather than creating redundant nodes. This loop refines the memory without uncontrolled expansion, enabling continual learning in application.

\paragraph{Local evolution} adapts the agent's working context to shifting UI states. Figure~\ref{fig:self_evolve} (b) highlights consistent performance improvements, such as a $\sim$15\% gain on Amazon. Empirically, the agent triggered Dynamic Relevance Assessment frequently (\emph{e.g.,} 39 times on Google Maps) to detect workflow ``phase shifts.'' For instance, in a task finding EV charging near museums, \textsc{HyMEM} successfully identified the transition from generic navigation to specific filtering, explicitly reasoning that the task phase had changed. This active context management maintains high relevance throughout long-horizon tasks, whereas the static baseline fails as context drifts from the initial retrieval.

\subsubsection{Impact of Memory Size}
\label{sec:memory_size}

To investigate the scaling behavior of our hybrid graph memory system, we analyze three aspects: (i) how performance improves with larger memory graphs; (ii) how adding more memory items influences success; and (iii) how efficiently our graph representation compresses redundant trajectories.

\paragraph{Size of Memory Graph} As shown in Table~\ref{tab:memory_scaling}, increasing the size of the memory graph leads to consistent gains in task success across domains. For instance, in the Amazon domain, scaling the memory from 500 to 1,000 trajectories boosts success from 19.5\% to 31.7\%, while using the largest graph (8,000 examples) achieves 63.4\%. This trend demonstrates that larger memory graphs enable more effective retrieval of relevant trajectories, enhancing the agent's performance.

\begin{figure}[h]
    \centering
    \includegraphics[width=\linewidth]{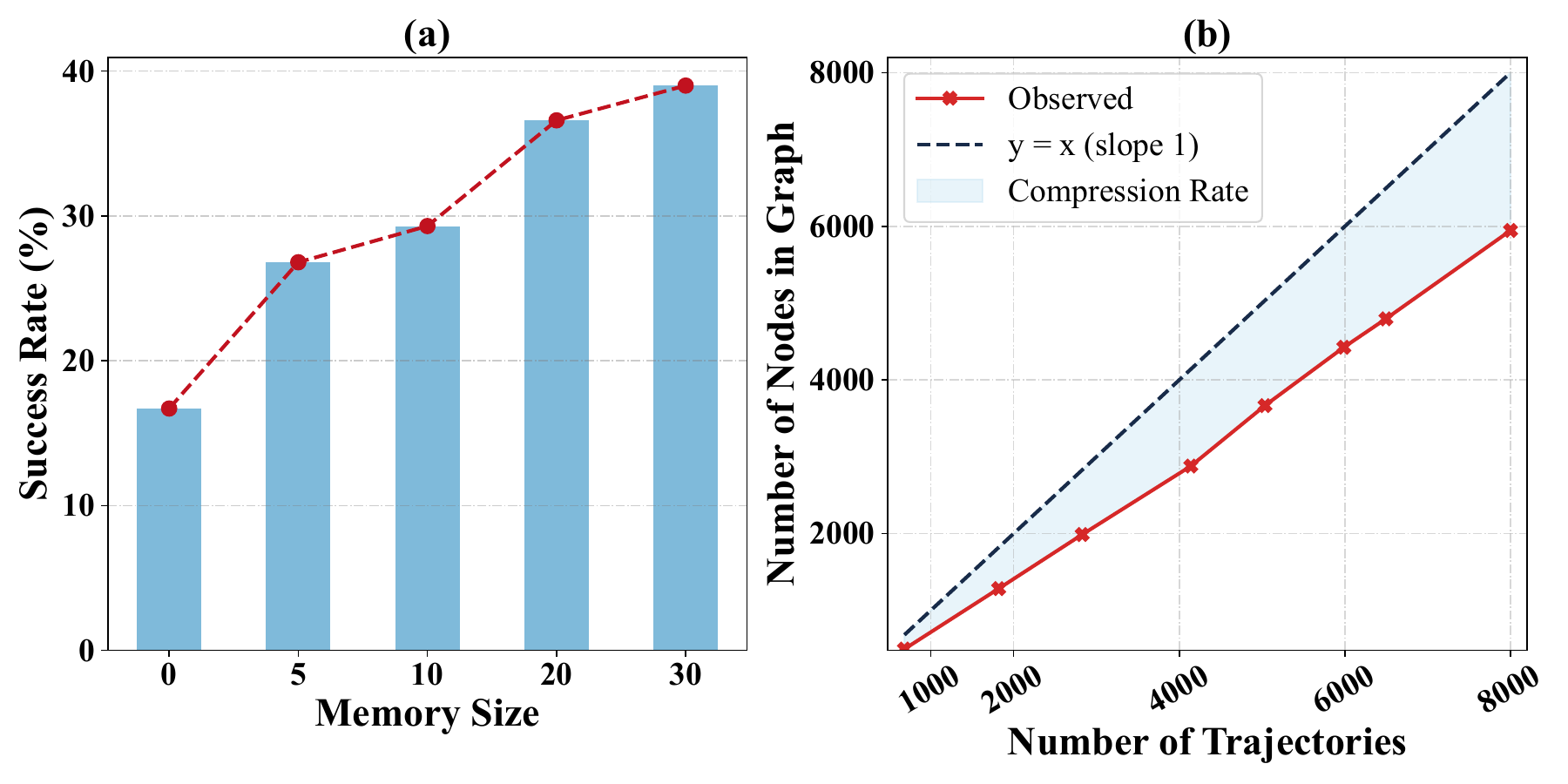}
    \caption{(a) Scaling Behavior of Memory Size in Coursera domain and (b) Graph Compression Analysis.}
    \label{fig:graph_size}
\end{figure}

\paragraph{Number of Retrieved Memory Items} Figure~\ref{fig:graph_size}(a) further illustrates this scaling trend by showing how success rate improves as we increase the number of retrieved memory items. We observe steady returns, suggesting that richer memory inputs contribute to better performance.


\paragraph{Graph Compression} Finally, as shown in Figure~\ref{fig:graph_size}(b), our graph memory exhibits a desirable compression property by merging semantically similar trajectories into unified nodes. As the number of raw trajectories increases, the number of graph nodes grows sublinearly, enabling efficient scaling without linear memory overhead.

\begin{table}[h]
\centering
\small
\setlength{\tabcolsep}{4pt} 
\begin{tabular}{lcccccc}
\toprule
\textbf{Memory Size} & \textbf{500} & \textbf{1000} & \textbf{2000} & \textbf{5000} & \textbf{8000}\\
\midrule
\textbf{Amazon}   & 19.5\% & 31.7\% & 63.4\% & 65.9\% & 63.4\% \\ 
\textbf{Google Map} & 31.7\% & 46.3\% & 51.2\%  & 51.2\% & 56.1\%   \\
\textbf{AllRecipes} & 11.1\%    & 13.3\%    & 13.3\% & 15.6\% & 20.0\%   \\
\bottomrule
\end{tabular}
\caption{
    Impact of Memory Bank Size on Task Success across different domains.
    Increasing the number of stored trajectories significantly improves the retrieval quality and overall agent success rate.
}
\label{tab:memory_scaling}
\end{table}

\subsubsection{Similarity and Diversity Trade-off}

As stated in Section \ref{method:retrieval}, graph-based retrieval and expansion can increase the diversity of retrieved trajectories. To investigate the trade-off between similarity and diversity in our graph-based memory retrieval, we compare three retrieval strategies: 
(i) $|\widetilde{\mathcal{N}}_{\text{initial}}| = 10$, where all 10 trajectories are retrieved solely using FAISS based on multimodal similarity to the current task, prioritizing similarity over diversity; 
(ii) $|\widetilde{\mathcal{N}}_{\text{initial}}| = 5$, where 5 seed trajectories are retrieved via FAISS and the remaining 5 are selected from their 1-hop graph neighbors, keeping the balance between similarity and diversity; and 
(iii) $|\widetilde{\mathcal{N}}_{\text{initial}}| = 1$, where only 1 similar trajectory is used as the seed and the other 9 are drawn from the 1-hop neighborhood, prioritizing diversity. 

As shown in Table~\ref{tab:similarity_diversity_tradeoff}, the strategy with $|\widetilde{\mathcal{N}}_{\text{initial}}|=5$ achieves the highest overall task success rate across all three domains (Amazon: 63.4\%, Coursera: 54.8\%, GMap: 53.7\%), outperforming the other two settings. This suggests that combining relevant yet diverse trajectories provides the most effective contextual grounding for the agent. It also highlights the importance of balancing semantic similarity and diversity in memory retrieval to support generalizable and effective reasoning.

\begin{table}[h]
\centering
\small
\setlength{\tabcolsep}{6pt}

\begin{tabular}{lccc}
\toprule
\textbf{Strategy} 
& \textbf{Amazon} 
& \textbf{Coursera} 
& \textbf{GMap} \\
\midrule
\textbf{$|\widetilde{\mathcal{N}}_{\text{initial}}|=5$}  & \textbf{63.4\%} & \textbf{54.8\%} & \textbf{53.7\%} \\
\textbf{$|\widetilde{\mathcal{N}}_{\text{initial}}|=10$} & 53.7\% & 31.0\% & 46.3\% \\
\textbf{$|\widetilde{\mathcal{N}}_{\text{initial}}|=1$} & 46.3\% & 31.0\% & 48.8\% \\
\bottomrule
\end{tabular}
\caption{
Task success rate under different retrieval strategies, showing the similarity and diversity trade-off.
}
\label{tab:similarity_diversity_tradeoff}
\end{table}




\section{Conclusion}
Existing GUI agent memory approaches remain limited by flat storage and retrieval that do not support structured organization or continual evolution. 
To solve it, inspired by human memory, we proposed Hybrid Self-evolving Structured Memory (\textsc{HyMEM}), which unifies discrete high-level symbolic strategy nodes with continuous trajectory embeddings in a graph-based memory. This structure enables (i) multi-hop, seed-and-expand retrieval for more informative and diverse context, (ii) self-evolution through node add/merge/replace updates as new trajectories arrive, and (iii) on-the-fly working-memory refreshing during inference to better support multi-turn execution and phase changes.
Extensive experiments across multiple benchmarks demonstrate that \textsc{HyMEM} consistently strengthens open-source GUI agents and substantially narrows—and in some settings surpasses—the gap to strong closed-source systems. 

Looking ahead, the hybrid, self-evolving memory paradigm offers a promising foundation for scalable agent improvement: extending the graph to richer strategy abstractions, improving update policies under distribution shift, and leveraging memory evolution for continual learning.

\section{Limitations}

While our proposed method demonstrates strong performance gains across multiple benchmarks, there are several limitations for further exploration. First, although the memory update mechanism leverages the base VLM to decide whether an update is necessary, this approach still relies on heuristics rather than being fully optimized. A more principled solution—such as training the model with reinforcement learning could enable the agent to learn when and how to use or update memory more effectively, potentially leading to further improvements. Second, due to limitations in computational resources, we were unable to evaluate our method on larger-scale models (\emph{e.g.,} 32B or 70B). Investigating the scalability of our approach with more powerful backbones remains an important direction for future work.

\section{Ethical considerations}
This work focuses on enhancing the memory and reasoning capabilities of open-source GUI agents. All experiments are conducted using publicly accessible websites and publicly available datasets that are commonly used in prior research. Our data collection and evaluation procedures strictly adhere to website usage policies and do not involve any login-protected content, private user data, or actions that would violate terms of service. As such, our work does not present any privacy risks or ethical concerns in its current form.

However, in real-world applications, deploying autonomous agents capable of interacting with web interfaces at scale necessitates careful safeguards. Without proper constraints, such agents could unintentionally perform harmful or unauthorized actions. We admit it is crucial to incorporate robust safety mechanisms, access controls, and ethical oversight when applying this technology beyond controlled research settings.

\bibliography{custom}

\clearpage
\appendix

\section{Implementation Details}
\label{sec:appendix_grounding}

\paragraph{Evaluation Benchmarks}
We evaluate our method on three public benchmarks.
We introduce our detailed settings here:
\begin{itemize}
    \item \textbf{WebVoyager}~\citep{he2024webvoyager}: A dataset focusing on multimodal reasoning in dynamic web environments. We show results on four representative websites in the main table \ref{tab:main_results}: Amazon (Amz), Coursera (Cour), AllRecipes (Recp), and Google Maps (Map). 

    \item \textbf{Multimodal-Mind2Web}~\citep{mind2web, zheng2024seeact_multimodal_mind2web}: A large-scale benchmark with 2,000+ open-ended tasks, which include domains and websites not seen during training, thereby evaluating the out-of-distribution (OOD) generalizability of the agent. We skip tasks where the associated website is no longer accessible at evaluation time.
    
    \item \textbf{MMInA}~\citep{tian2025mminabenchmarkingmultihopmultimodal}: A real-world multistep benchmark requiring multimodal reasoning and planning across domains such as shopping and travel. We evaluate on the Wikipedia domain using the first 100 tasks.
    
\end{itemize}
\paragraph{Continuous Encoder Training}
\label{appendix:training_recipe}
For training data, we use 1,009 trajectories from the Mind2Web training set \cite{mind2web}, along with 1,000 successful trajectories collected through rollouts by our agent. These tasks are generated from publicly accessible websites such as Amazon, Coursera, Google Maps, and others.

Using this data, we perform parameter-efficient fine-tuning on the Q-Former and LoRA layers within the VLM encoder. To improve efficiency, we apply LoRA with a rank of 16 and share parameters across all layers of the Q-Former. As a result, only 1.2\% of the total model parameters are updated during training. These parameter- and data-efficient design choices allow the entire training process to be completed within 20 hours on a single NVIDIA H100 GPU.

Empirically, we observe that training converges rapidly, and a single epoch is sufficient to achieve strong performance. We apply the same training strategy across multiple vision-language models, including Qwen2.5-VL-7B \cite{bai2025qwen25vltechnicalreport}, Qwen3-VL-8B \cite{bai2025qwen3vltechnicalreport}, and UI-TARS-1.5-7B \cite{qin2025uitarspioneeringautomatedgui}.

\section{Additional Qualitative Results}
\label{sec:appendix_results}

\paragraph{Visualization of Hybrid Memory}

\begin{figure}[h]
    \centering
    \includegraphics[width=\linewidth]{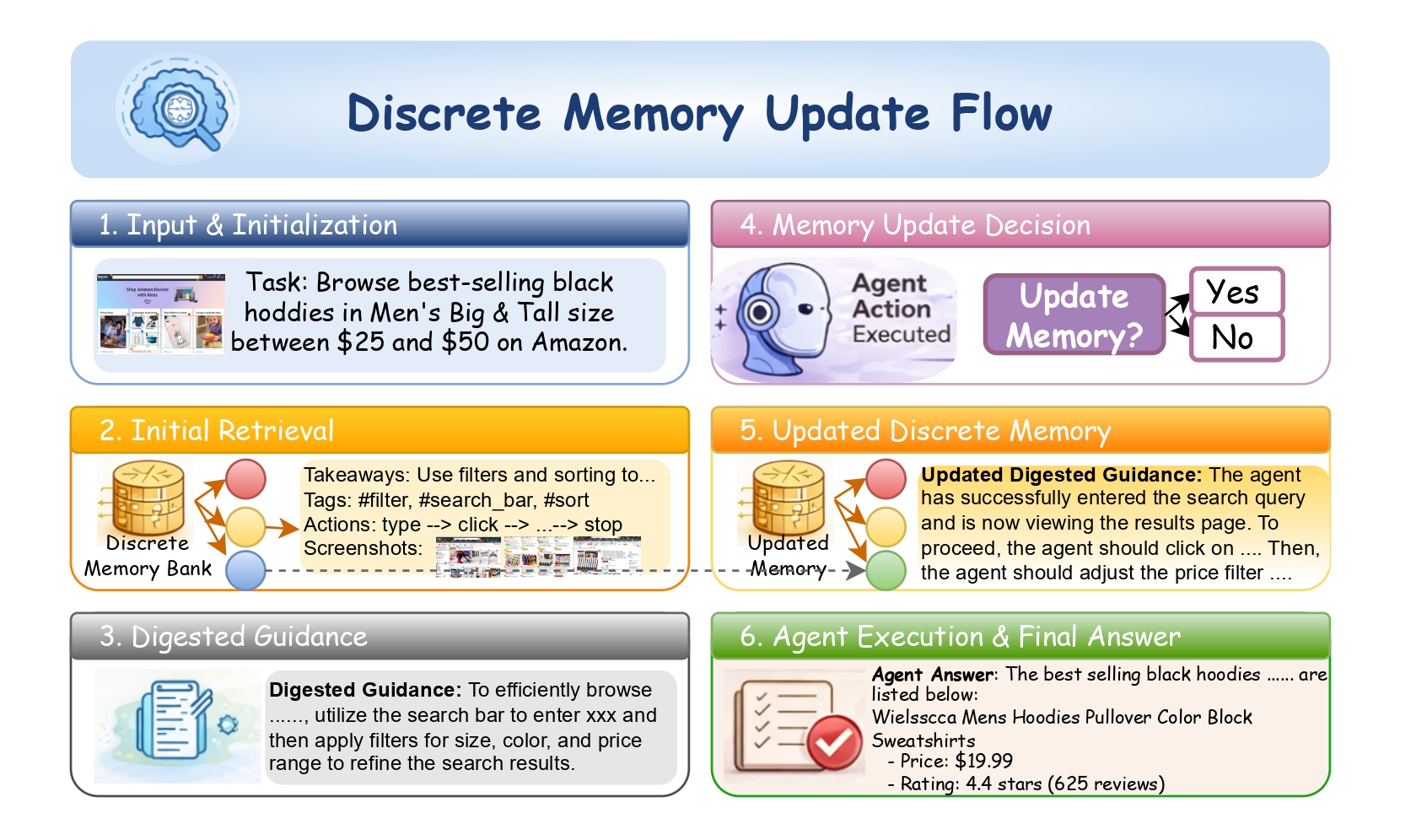}
    \caption{Visualization of Discrete Memory Update Flow.
    }
    \label{fig:discrete_memory}
\end{figure}

Figure~\ref{fig:discrete_memory} illustrates the complete workflow of discrete memory within our hybrid memory system, highlighting how it is generated, updated, and utilized during agent execution. The process begins with user task input and environment initialization (Step 1), followed by the retrieval of relevant past memory from the discrete memory bank (Step 2). This retrieval includes structured takeaways, action traces, and visual references, which are then distilled into actionable guidance (Step 3).

As the agent executes actions in the environment, a memory update decision is triggered (Step 4). If new knowledge or behaviors are observed, the  memory is updated accordingly (Step 5), incorporating the latest digested guidance. Finally, the agent uses this refined memory to guide subsequent actions and generate the final answer (Step 6). This workflow enables the agent to continually improve its decision-making by leveraging structured, interpretable memory updates throughout the task.

\begin{figure}[h]
    \centering
    \includegraphics[width=0.8\linewidth]{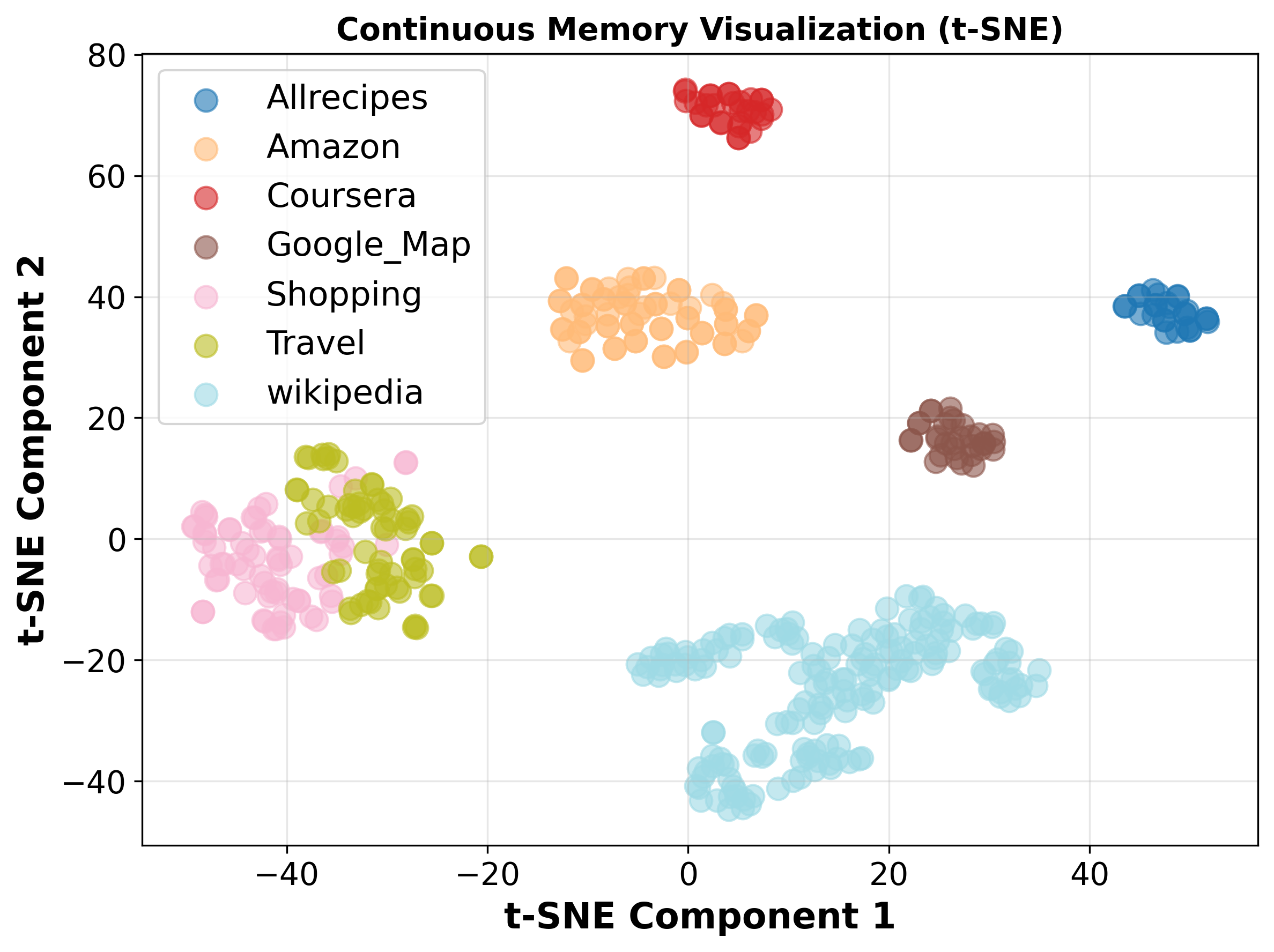}
    \caption{
    t-SNE visualization of continuous memory embeddings across domains. 
    Each point represents a stored trajectory embedding, color-coded by domain.
    }
    \label{fig:tsne_memory}
\end{figure}

To better understand how hybrid memory organizes knowledge across domains, 
we visualize the continuous memory embeddings using t-SNE, as shown in Figure~\ref{fig:tsne_memory}. 
Each point represents a trajectory stored in memory, and colors indicate different domains. 
The clear clustering structure demonstrates that the memory embeddings capture meaningful semantic distinctions between domains. 
For instance, trajectories from Amazon, Coursera, and Google Map form well-separated clusters, 
while related domains such as Travel and Shopping show some proximity, reflecting shared behavioral patterns. 
This interpretable structure supports effective cross-domain retrieval and highlights the advantage of learning unified memory representations.

\paragraph{Case Study}

In this example~\ref{fig:hybrid_memory_case_study}, the agent is tasked with finding a USB-C hub on Amazon that meets some specific criteria. The agent powered by Qwen2.5-VL-7B + Hybrid Memory successfully completes the task by leveraging both current context and retrieved memory.

The agent first interprets the goal accurately, identifying key constraints such as compatibility, port requirements, and price. During execution, it recognizes the need to update memory and retrieves relevant trajectories involving similar filtering and product selection behavior. These include both high-level concepts (\emph{e.g.,} filtering by features) and low-level UI actions (\emph{e.g.,} selecting “Best Sellers”). Guided by this contextualized memory, the agent efficiently navigates the interface, applies the correct filters, and selects a product that meets all constraints. This case illustrates how hybrid memory enhances reasoning by grounding decisions in both semantic and procedural knowledge.

In contrast, the same task is attempted using the Qwen2.5-VL-7B base model without access to hybrid memory~\ref{fig:base_model_case_study}. Initially, the agent is able to locate the search bar and input the query correctly. It also selects the sorting option, but fails to apply the necessary filters: Best Sellers. As the task progresses, the model repeatedly opens similar product pages without making meaningful progress toward the goal. Eventually, it hits the maximum step limit without identifying a suitable product. Without access to relevant examples or structured memory, the model struggles to decompose the high-level instruction into effective UI actions and loses context part way through the task.

\section{Prompts Used in \textsc{HyMEM}}
\label{sec:appendix_prompts}

\paragraph{Guidance Digestion Prompt}
This prompt is used in \textsc{HyMEM}'s inference-time digestion stage to convert retrieved experience summaries into a compact \emph{Guidance Block}.
As shown in Figure~\ref{fig:prompt_guidance_digestion}, given the current task intent, current screenshot, and retrieved summaries, it synthesizes task-specific navigation cues (\emph{e.g.,} key actions/filters) while explicitly avoiding termination-related instructions.

\paragraph{Local Self-Evolution Prompt}
This prompt is used in \textsc{HyMEM}'s inference-time \emph{local evolution} stage to decide whether the current working memory should be refreshed after each environment transition.
As shown in Figure~\ref{fig:prompt_local_evolve}, it conservatively distinguishes \emph{operational errors} (no refresh) from genuine \emph{phase shifts} (refresh), and identifies which existing takeaways should be preserved when re-retrieving for the new phase.

\paragraph{Global Evolution Prompt}
This prompt is used in \textsc{HyMEM}'s \emph{global evolution} stage to deduplicate newly observed successful trajectories against existing strategy nodes and decide whether to \textsc{ADD}, \textsc{UPDATE}, or \textsc{REPLACE} a node.
As illustrated in Figure~\ref{fig:prompt_global_evolution}, the judge leverages both textual metadata (task/takeaway/tags/domain) and visual context (screenshots) and outputs a strict JSON decision for downstream graph updates.

\section{LLM Usage Statement}
Large Language Models (LLMs) were utilized in this work solely for the purpose of grammatical refinement and language polishing to enhance readability. No substantive edits were made to the scientific content, conceptual development, or experimental results by these tools. The authors retain full responsibility for the final content and accuracy of the manuscript.

\clearpage

\begin{figure*}[h]
    \centering
    \includegraphics[width=\linewidth]{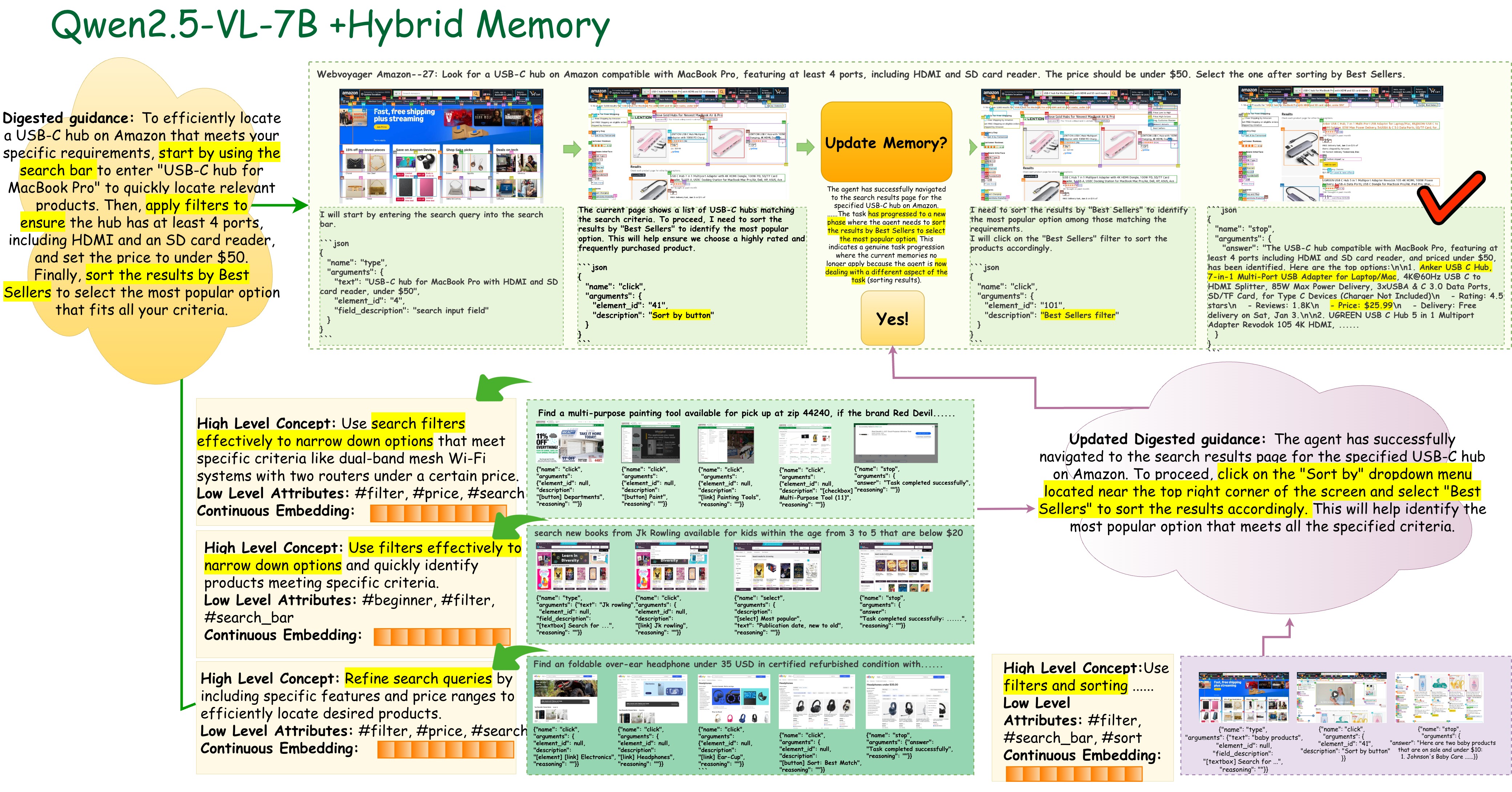}
    \caption{Successful case by Qwen2.5-VL-7B + Hybrid Memory in the Amazon task. The model
accurately understands the goal, retrieves relevant memory, and selects the correct product by filtering.
    }
    \label{fig:hybrid_memory_case_study}
\end{figure*}

\begin{figure*}[h]
    \centering
    \includegraphics[width=\linewidth]{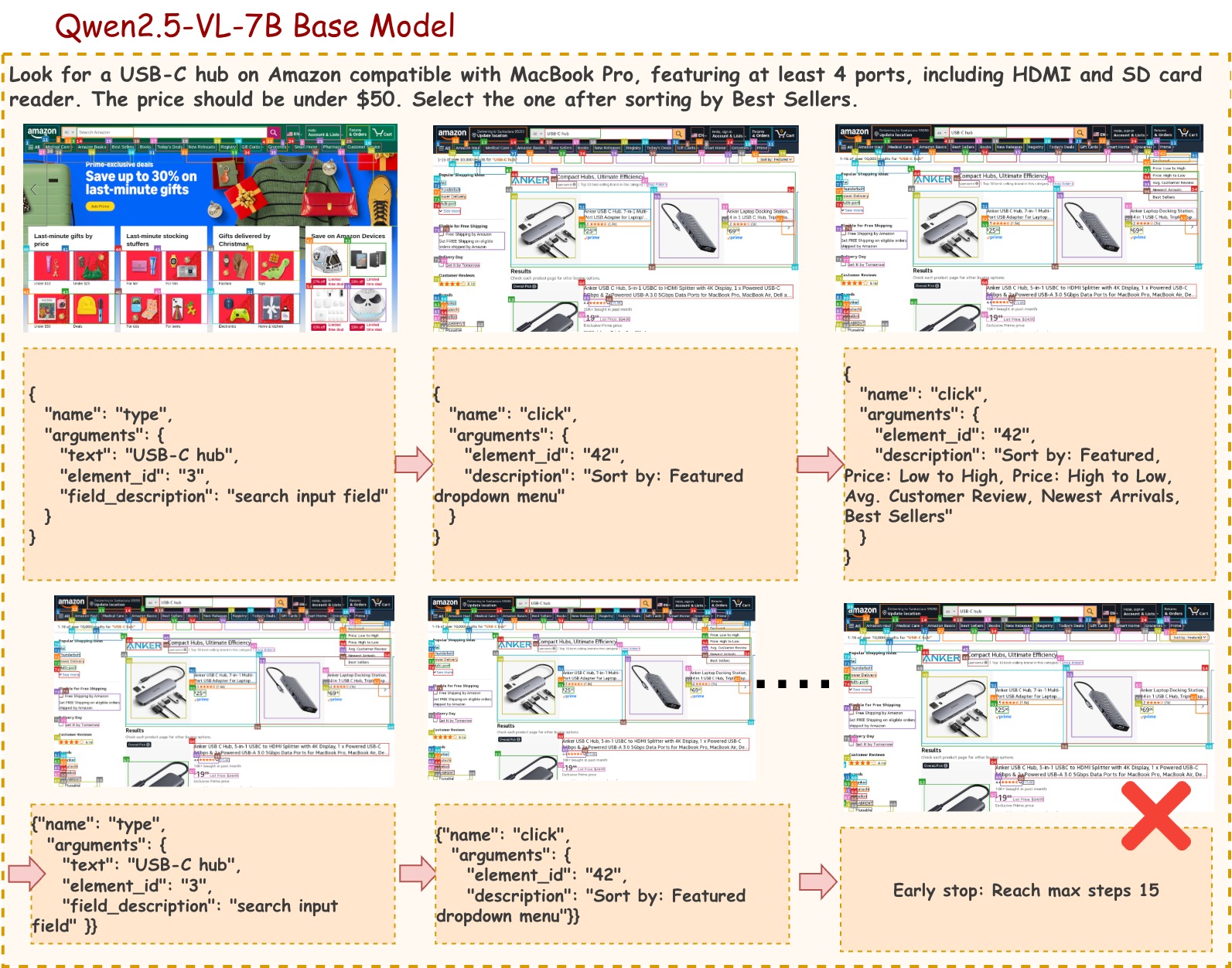}
    \caption{Failure case by Qwen2.5-VL-7B base model in the Amazon task. The model does not know how to filter products by the requirement, and loses track of the task state.
    }
    \label{fig:base_model_case_study}
\end{figure*}

\clearpage

\begin{figure*}[h]
    \centering
    \includegraphics[width=\linewidth]{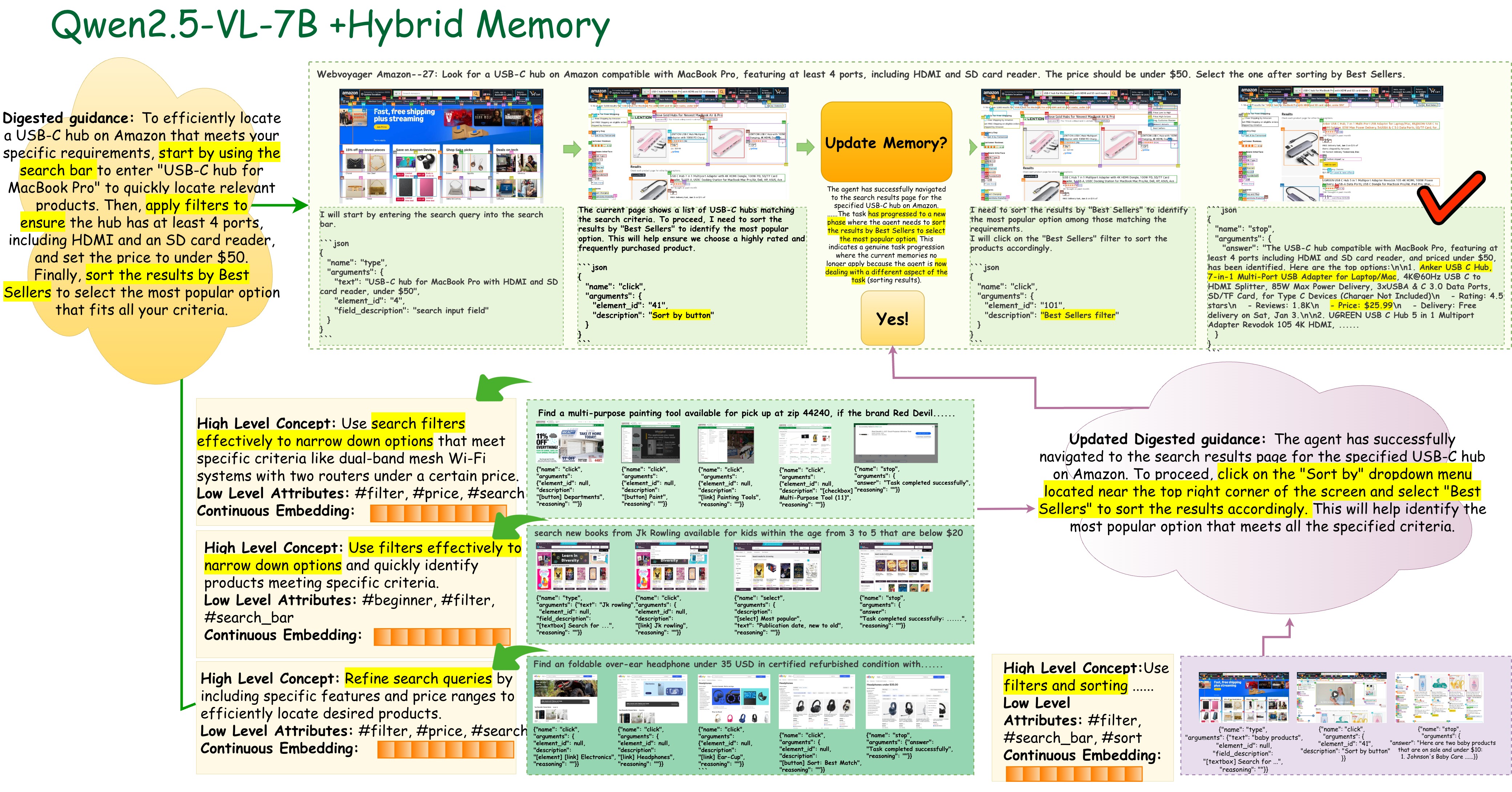}
    \caption{Successful case by Qwen2.5-VL-7B + Hybrid Memory in the Amazon task. The model
accurately understands the goal, retrieves relevant memory, and selects the correct product by filtering.
    }
    \label{fig:hybrid_memory_case_study}
\end{figure*}

\begin{figure*}[h]
    \centering
    \includegraphics[width=\linewidth]{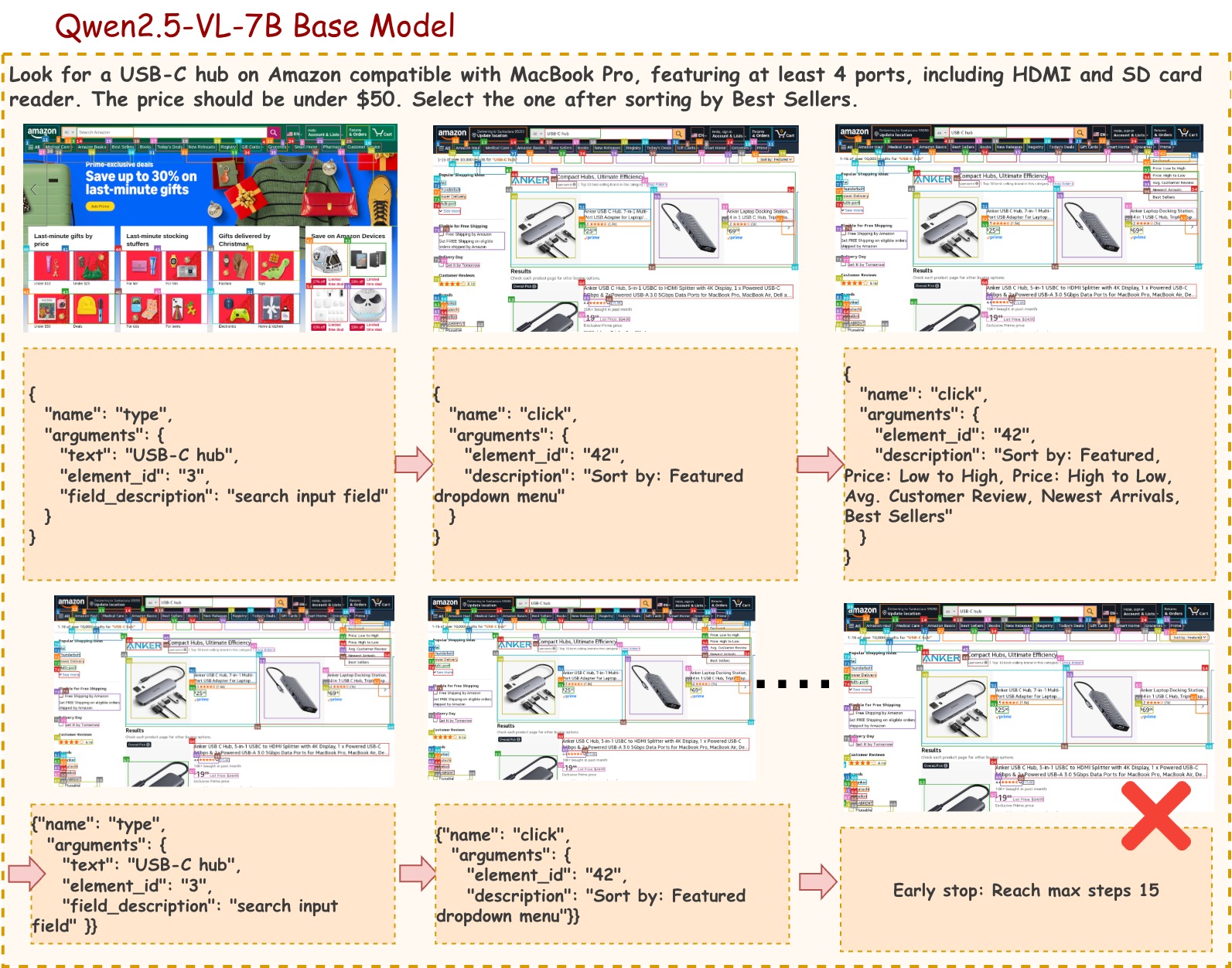}
    \caption{Failure case by Qwen2.5-VL-7B base model in the Amazon task. The model does not know how to filter products by the requirement, and loses track of the task state.
    }
    \label{fig:base_model_case_study}
\end{figure*}

\clearpage

\begin{figure*}[t]
\centering
\begin{minipage}{\textwidth}
\begin{PromptBox}{Guidance Digestion Prompt}

\PromptSection{System Instruction}
\begin{lstlisting}[style=promptstyle]
You are an expert at analyzing past GUI agent experiences to help with a new task.
Given the current task, current screenshot, and retrieved experience summaries,
synthesize them into focused, actionable guidance.

Output format: ONE concise paragraph (2-3 sentences) that answers:
1. Which strategies from past experiences are MOST relevant to this specific task?
2. What key actions or filters should be prioritized?

IMPORTANT RULES:
- Focus ONLY on navigation/search strategies, NOT on when to stop.
- Do NOT mention stopping, completing, or finishing the task.
- Do NOT give instructions about providing answers or explanations.
- Be specific to the current task. Do NOT just repeat the summaries.
- Do NOT use bullet points. Write as a coherent paragraph.
\end{lstlisting}

\PromptSection{Input Prompt}
\begin{lstlisting}[style=promptstyle]
Current task:
{TASK_INTENT}

Current screenshot:
{SCREENSHOT}

Retrieved experience summaries:
{RETRIEVED_SUMMARIES}
\end{lstlisting}
\end{PromptBox}
\end{minipage}
\caption{\textbf{Guidance Digestion Prompt} used to synthesize retrieved experience summaries into a compact \emph{Guidance Block} for inference-time navigation.}
\label{fig:prompt_guidance_digestion}
\end{figure*}

\clearpage

\begin{figure*}[t]
\centering
\begin{minipage}{\textwidth}
\begin{PromptBox}[colframe=DeepGreen, colback=LightGreenBg]{Local Self-Evolution Prompt}

\PromptSection[DeepGreen]{System Instruction}
\begin{lstlisting}[style=promptstyle]
You are evaluating whether the retrieved experience memories are still relevant for the task.

KEY PRINCIPLES:
1. Memories provide NAVIGATION STRATEGIES (how to search, filter, click, scroll, etc.), NOT answers.
2. Be VERY CONSERVATIVE - default to keeping current memories.
3. CRITICAL: Distinguish between OPERATIONAL ERRORS vs MEMORY RELEVANCE:

   OPERATIONAL ERRORS (DO NOT update memory):
   - Agent landed on wrong page (login page, unrelated website, error page)
   - Agent is stuck or action failed
   - Agent made a navigation mistake
   -> These are execution errors, NOT memory problems. The current memories are still valid for when agent gets back on track.

   MEMORY RELEVANCE ISSUE (May update memory):
   - Task has genuinely progressed to a completely new phase
   - Agent successfully completed initial steps and now needs different strategies
   - Current screen shows the task is in a fundamentally different domain than memories cover
   -> Only update if memories are irrelevant to the ACTUAL TASK PROGRESS, not temporary navigation mistakes.

4. PRESERVE USEFUL MEMORIES: When updating, some existing takeaways may still be valuable:
   - General domain knowledge that applies across task phases
   - Strategies that might be needed again later (e.g., navigation, filtering)
   - Takeaways that provide complementary information to new phase requirements

Your job: Decide if memories should be updated, but ONLY for genuine relevance issues, NOT operational errors.
When updating, also identify which existing takeaways should be PRESERVED and combined with new retrieval.

OUTPUT FORMAT (strict):
Decision: KEEP or UPDATE
Preserve: [comma-separated IDs of takeaways to keep, or NONE]
Reason: one sentence (no more than 20 words)
\end{lstlisting}

\PromptSection[DeepGreen]{Input Prompt}
\begin{lstlisting}[style=promptstyle]
Current task:
{TASK_INTENT}

Previous screenshot:
{SCREENSHOT_t}

Current screenshot:
{SCREENSHOT_t+1}

Current guidance block (optional):
{GUIDANCE_BLOCK}

Current retrieved takeaways (with IDs):
{TAKEAWAYS_WITH_IDS}
\end{lstlisting}
\end{PromptBox}
\end{minipage}
\caption{\textbf{Local self-evolution prompt} for conservative working-memory refresh: it updates only on genuine phase shifts and preserves reusable takeaways.}
\label{fig:prompt_local_evolve}
\end{figure*}

\clearpage
\begin{figure*}[p] 
\centering
\begin{minipage}{\textwidth}
\begin{PromptBoxFloat}[colframe=DeepOrange, colback=LightOrangeBg]{Global Evolution Prompt}

\PromptSection[DeepOrange]{System Instruction}
\begin{lstlisting}[style=promptstyle]
You deduplicate GUI agent strategy takeaways using both text and visual context.
Output valid JSON only, no markdown code blocks.
\end{lstlisting}

\PromptSection[DeepOrange]{User Prompt}
\begin{lstlisting}[style=promptstyle]
You are managing a knowledge base of GUI agent strategies for deduplication.

NEW TRAJECTORY (see Screenshot 1):
- task_description: {new_task}
- takeaway: {new_takeaway}
- tags: {new_tags}
- domain: {new_domain}

EXISTING SIMILAR TRAJECTORIES:
1. ID: {traj_id} (similarity: {sim_score:.3f}, see {screenshot_ref})
   task_description: {existing_task}
   takeaway: {existing_takeaway}
   tags: {existing_tags}
   domain: {existing_domain}
[... more similar trajectories ...]

DECIDE one of:
1. UPDATE: NEW describes the SAME or very similar strategy as an existing trajectory.
   -> Keep the existing node but IMPROVE its takeaway by incorporating insights from NEW.
   -> Provide the improved takeaway (must start with "takeaway:").
   -> Provide the complete updated tags list (can add, remove, or keep tags).

2. REPLACE: NEW describes the SAME strategy as an existing trajectory, but NEW is STRICTLY BETTER
   (more specific, more actionable, more complete trajectory).
   -> Remove the existing node, add NEW as replacement.

3. ADD: NEW describes a GENUINELY DIFFERENT strategy not covered by existing trajectories.
   -> Add NEW as a new node.

Use the screenshots to understand the visual context and UI state of each trajectory.

OUTPUT (JSON only, no markdown):
{
  "action": "update" | "replace" | "add",
  "reasoning": "one sentence explanation",
  "target_id": "ID to update (required for update)",
  "updated_takeaway": "takeaway: improved takeaway text (required for update)",
  "updated_tags": ["#tag1", "#tag2", ...],
  "old_id": "ID to replace (required for replace)"
}
\end{lstlisting}

\end{PromptBoxFloat}
\end{minipage}
\caption{\textbf{Global evolution prompt} for strategy deduplication and graph maintenance. The judge selects \textsc{ADD}/\textsc{UPDATE}/\textsc{REPLACE} using multimodal context and returns a strict JSON decision for deterministic memory updates.}
\label{fig:prompt_global_evolution}
\end{figure*}

\end{document}